\documentclass[sigconf]{acmart}

\copyrightyear{2021}
\acmYear{2021}
\setcopyright{acmlicensed}\acmConference[CIKM '21]{Proceedings of the 30th ACM International Conference on Information and Knowledge Management}{November 1--5, 2021}{Virtual Event, QLD, Australia}
\acmBooktitle{Proceedings of the 30th ACM International Conference on Information and Knowledge Management (CIKM '21), November 1--5, 2021, Virtual Event, QLD, Australia}
\acmPrice{15.00}
\acmDOI{10.1145/3459637.3482460}
\acmISBN{978-1-4503-8446-9/21/11}

\usepackage{amsmath,amsfonts}
\usepackage{algorithmic}
\usepackage{graphicx}
\usepackage{textcomp}
\usepackage{xcolor}
\usepackage{url}
\usepackage{booktabs} 
\usepackage{subfigure}
\usepackage{mathrsfs}
\usepackage[ruled,vlined]{algorithm2e}
\usepackage[utf8]{inputenc}
\usepackage{array}
\usepackage{multirow}
\usepackage{makecell}
\usepackage{tabu}
\usepackage{enumitem}
\usepackage{balance}
\usepackage[normalem]{ulem}
\useunder{\uline}{\ul}{}
\newtheorem{mdefinition}{Definition}
\newtheorem{problem}{Problem}

\newcommand{\OurModel}{\emph{PSRNet} }
\newcommand{\OurModelNoSpace}{\emph{PSRNet}}
\newcommand{\OurBaseSRModel}{\emph{SNet} }
\newcommand{\OurBaseSRModelNoSpace}{\emph{SNet}}
\newcommand{\OurTSRModel}{\emph{TNet} }
\newcommand{\OurTSRModelNoSpace}{\emph{TNet}}
\newcommand{\OurSRModel}{\emph{STNet} }
\newcommand{\OurSRModelNoSpace}{\emph{STNet}}
\newcommand{\OurGANModel}{\emph{PGNet} }
\newcommand{\OurGANModelNoSpace}{\emph{PGNet}}

\newcommand{\tsc}[1]{\textsuperscript{#1}}

\begin{document}
\title{One-shot Transfer Learning for Population Mapping}

\author{
    Erzhuo Shao$\ast$\tsc{1,2}, 
    Jie Feng$\ast$\tsc{1,2}
    Yingheng Wang\tsc{2}, 
    Tong Xia\tsc{1,2}, 
    Yong Li$\dagger$\tsc{1,2}
}
\affiliation{%
  \institution{
  \tsc{1}Beijing National Research Center for Information Science and Technology (BNRist)}
}
\affiliation{%
    \institution{
        \tsc{2}Department of Electronic Engineering, Tsinghua University, Beijing, China, 100084}
    \country{}
}
\email{shaoerzhuo@gmail.com, fengj12ee@hotmail.com, wangyh20@mails.tsinghua.edu.cn} \email{xia-t17@tsinghua.org.cn, liyong07@tsinghua.edu.cn}

\thanks{$\ast$ Erzhuo Shao and Jie Feng contributed equally to this research.}
\thanks{$\dagger$ Yong Li is the corresponding author.}

\renewcommand{\shortauthors}{Erzhuo Shao \& Jie Feng, et al.}

\begin{abstract}
Fine-grained population distribution data is of great importance for many applications, \emph{e.g.}, urban planning, traffic scheduling, epidemic modeling, and risk control. However, due to the limitations of data collection, including infrastructure density, user privacy, and business security, such fine-grained data is hard to collect and usually, only coarse-grained data is available. Thus, obtaining fine-grained population distribution from coarse-grained distribution becomes an important problem. To tackle this problem, existing methods mainly rely on sufficient fine-grained ground truth for training, which is not often available for the majority of cities. That limits the applications of these methods and brings the necessity to transfer knowledge between data-sufficient source cities to data-scarce target cities. 

In knowledge transfer scenario, we employ single reference fine-grained ground truth in target city, which is easy to obtain via remote sensing or questionnaire, as the ground truth to inform the large-scale urban structure and support the knowledge transfer in target city. By this approach, we transform the fine-grained population mapping problem into a one-shot transfer learning problem.

In this paper, we propose a novel one-shot transfer learning framework \OurModel to transfer spatial-temporal knowledge across cities from three views. \emph{From the view of network structure}, we build a dense connection-based population mapping network with temporal feature enhancement to capture the complicated spatial-temporal correlation between population distributions of different granularities. \emph{From the view of data}, we design a generative model to synthesize fine-grained population samples with POI distribution and the single fine-grained ground truth in data-scarce target city. \emph{From the view of optimization}, after combining above structure and data, we propose a pixel-level adversarial domain adaption mechanism for universal feature extraction and knowledge transfer during training with scarce ground truth for supervision. 

Experiments on real-life datasets of $4$ cities demonstrate that \OurModel has significant advantages over $8$ state-of-the-art baselines by reducing RMSE and MAE by more than $25\%$. Our code and datasets are released in
\href{https://github.com/erzhuoshao/PSRNet-CIKM}{\textcolor{blue}{Github}}.
\end{abstract}

\ccsdesc[500]{Computing methodologies~Neural networks}
\ccsdesc[500]{Computing methodologies~Supervised learning by regression}
\ccsdesc[500]{Computing methodologies~Artificial intelligence}
\keywords{Transfer Learning, Population Distribution, Super-resolution}

\maketitle

\section{Introduction}

\begin{figure}[htbp]
    \centering
    \vspace{-0.5cm}
    \subfigure[Coarse-grained Population (CITY1).]{
        \includegraphics*[width=0.22\textwidth]{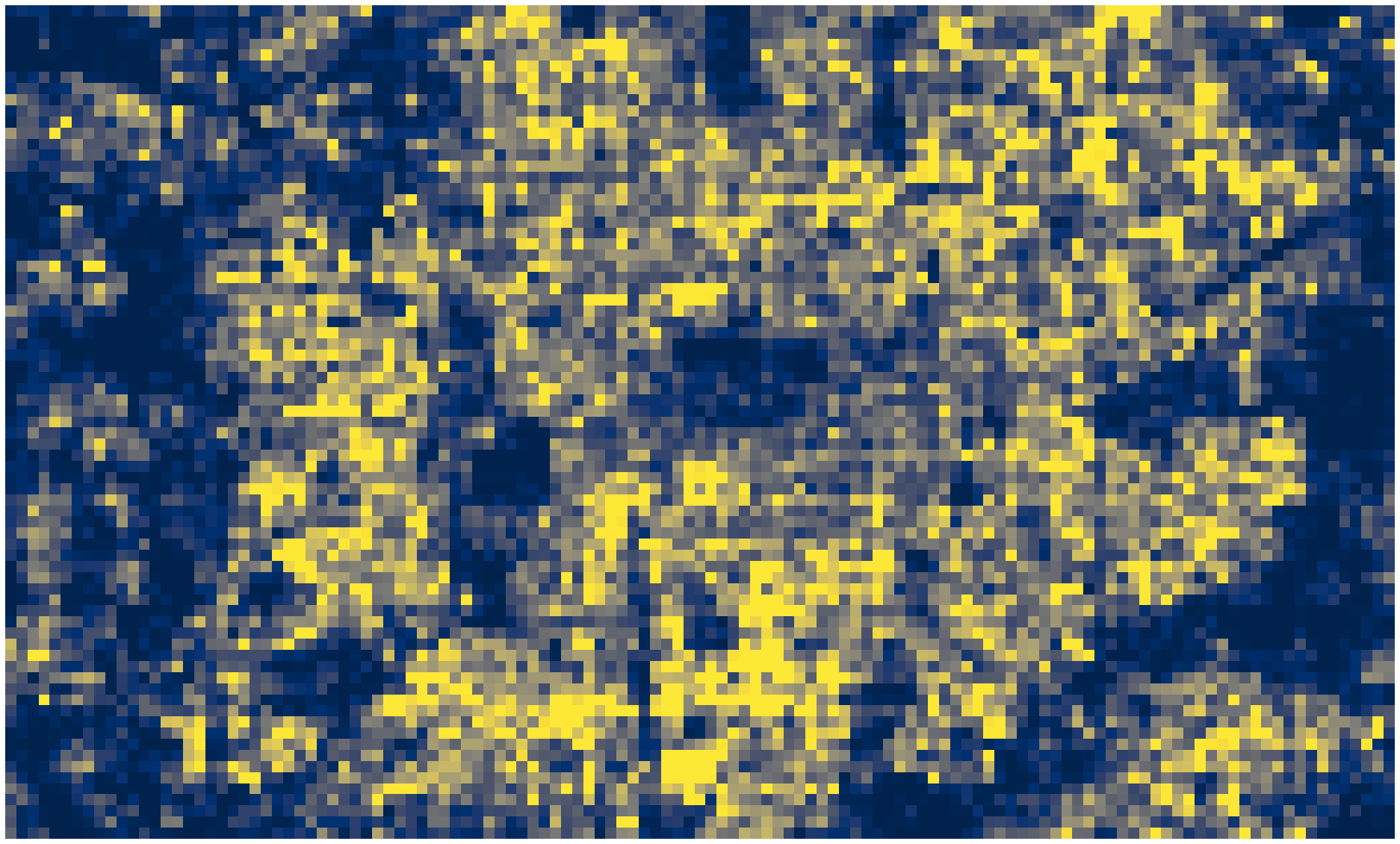}}
    \subfigure[Fine-grained Population (CITY1 $\times$ 4).]{
        \includegraphics*[width=0.22\textwidth]{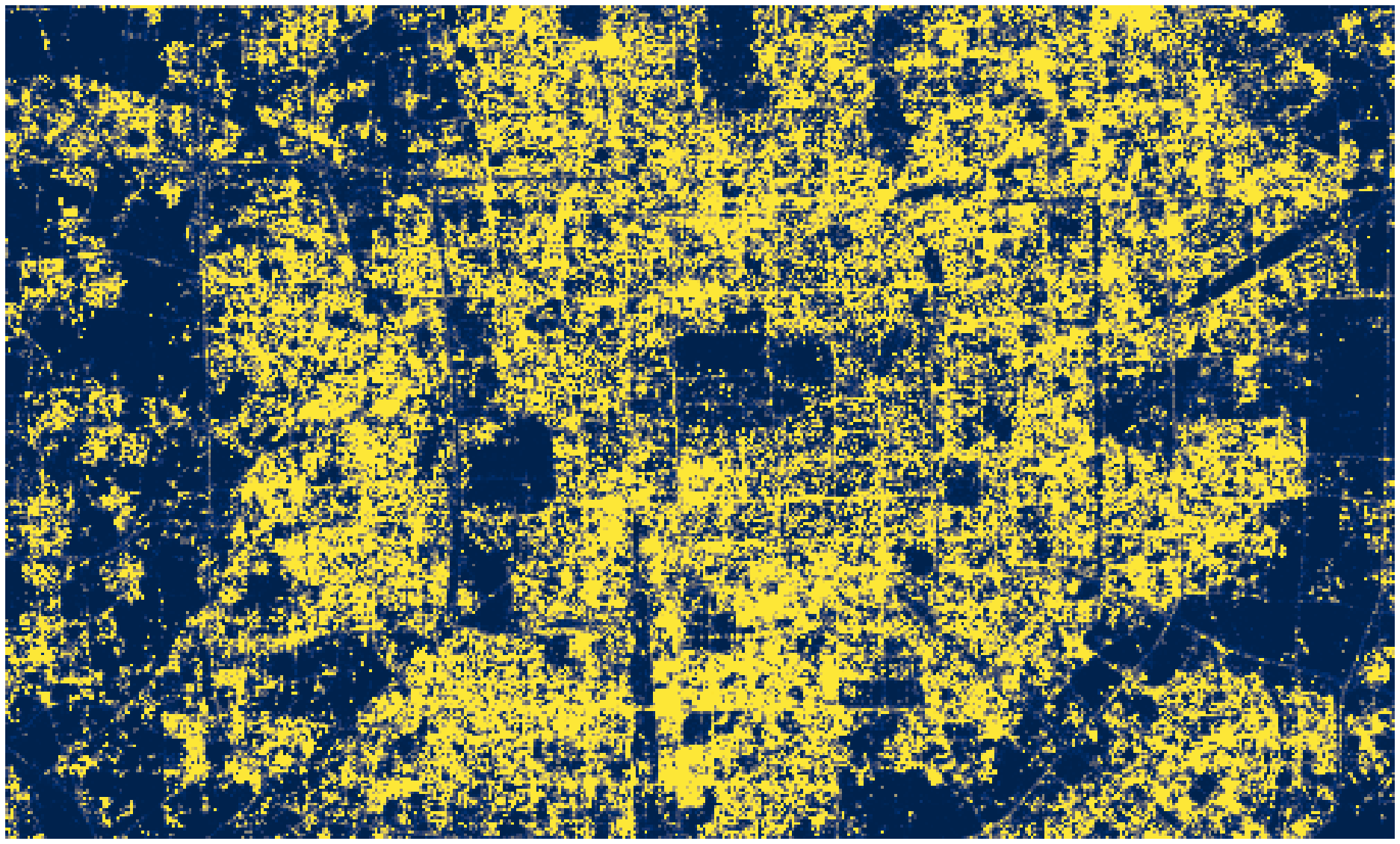}}
    \\\vspace{-0.2cm}
    \subfigure[Coarse-grained Population (CITY2).]{
        \includegraphics*[width=0.22\textwidth]{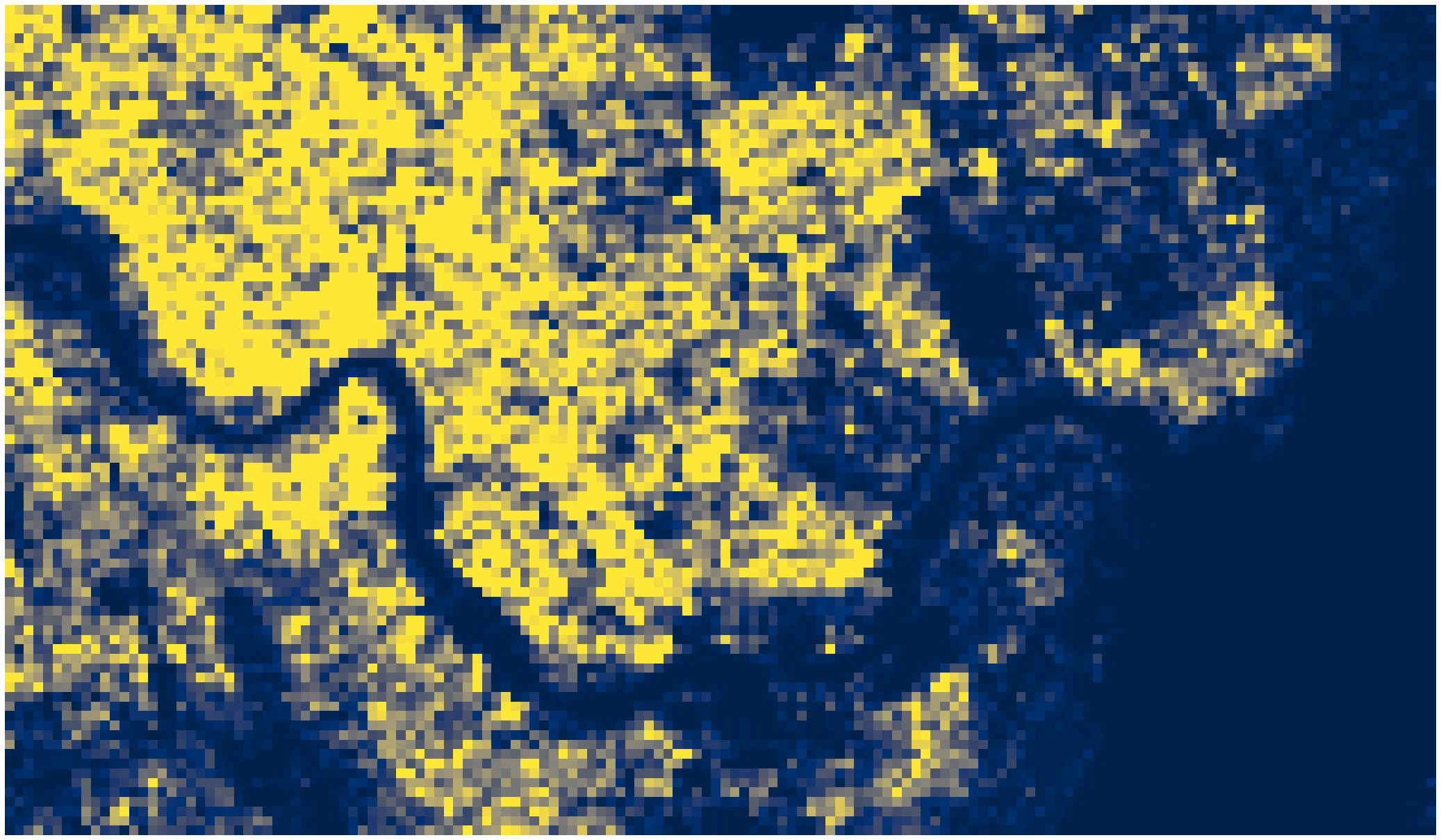}}
    \subfigure[Fine-grained Population (CITY2 $\times$ 4).]{
        \includegraphics*[width=0.22\textwidth]{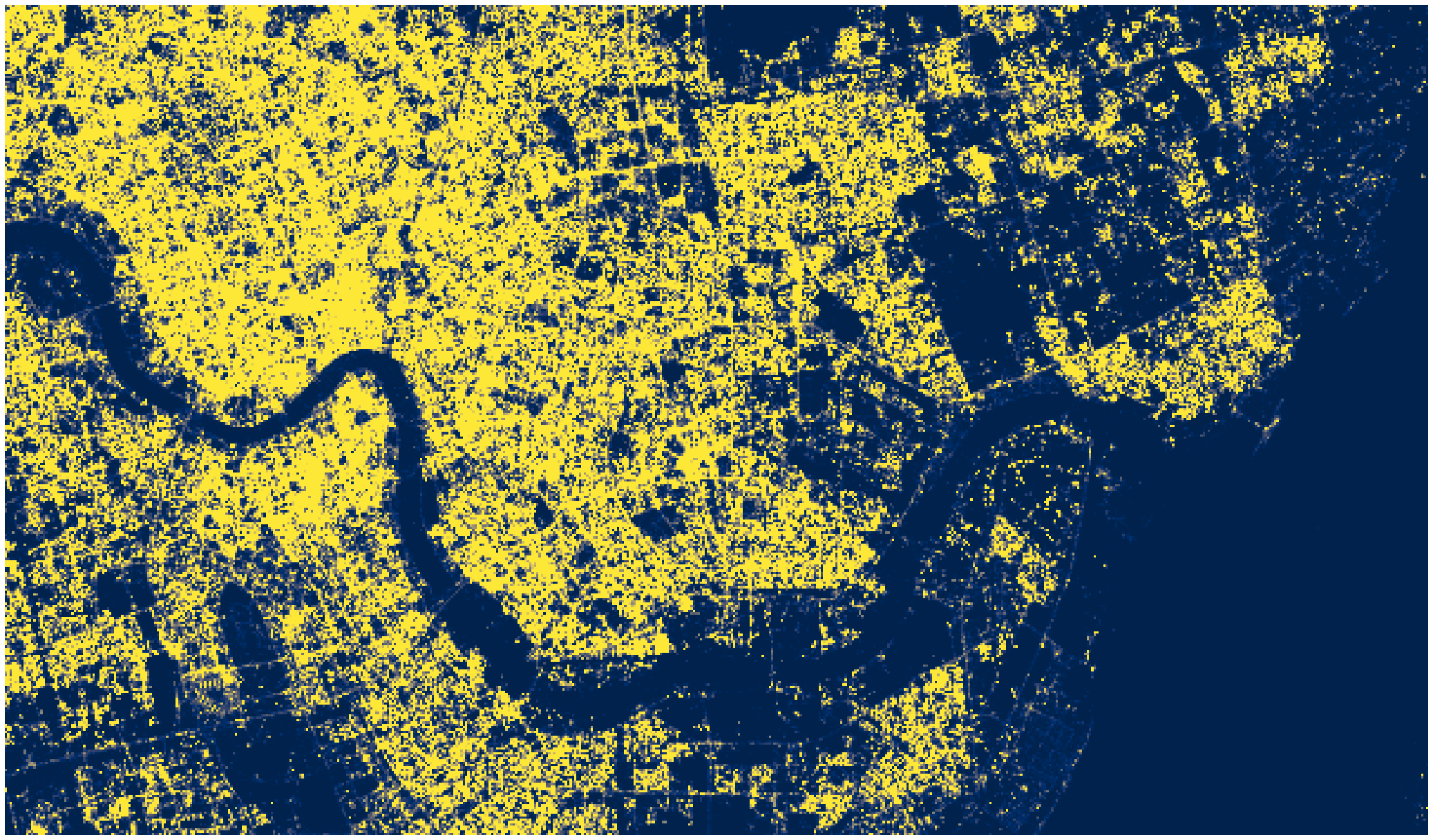}}
    \vspace{-0.5cm}
    \caption{Coarse-grained and fine-grained population distribution in CITY1 and CITY2. Lighter places have higher population.}
    \vspace{-0.4cm}
    \label{fig:intro}
\end{figure}

Fine-grained urban population distribution, \emph{e.g.}, the real-time population in $100m\times100m$ grids in the city, is of great importance for many applications, including urban planning, epidemic modeling, and transportation management. For example, with a large-scale data recording dynamic fine-grained population distribution, governments can make timely and effective policies about infrastructure construction for commute and public health during the pandemic disease. However, due to the limitations of current data collection systems along with the issues of user privacy and business security, such fine-grained population distribution is difficult to obtain and rarely to be open. Usually, only coarse-grained population distribution like $2km\times2km$ is obtainable. Thus, as Figure~\ref{fig:intro}, inferring fine-grained urban population distribution from coarse-grained data, also known as fine-grained population mapping problem~\cite{deville2014dynamic, stevens2015disaggregating}, becomes an important task.

Although existing neural network-based population mapping methods~\cite{zong2019deepdpm, liang2019urbanfm} have promising performance, these methods always rely on sufficient fine-grained population distribution as ground truth to supervise their training, which severely restricts their applications since sufficient fine-grained population data are usually not obtainable. 
Therefore, a method, which could infer fine-grained population distribution without sufficient fine-grained ground truth, would significantly broaden the applications of population mapping in data-scarce cities.
Fortunately, we could obtain prior knowledge of these cities from several approaches. First, we could extract universal and transferable knowledge about the relationship between coarse-grained population and fine-grained population from data-sufficient source city to support the population mapping in data-scarce target city. Second, although sufficient fine-grained population distribution data is unobtainable, a single static reference fine-grained ground truth is still easy-to-obtain by remote sensing~\cite{stevens2015disaggregating} or questionnaire. Third, POI (Point of Interest) distribution characterizes the large-scale urban structures and region functions. Its strong relationship with population and crowd flow makes it informative.

Thus, we transform the population inference problem into a one-shot transfer learning problem, since there is only one target domain's fine-grained ground truth in our scenario. We need to extract transferable knowledge from the data-sufficient source domain (city) and transfer it into data-scarce target domain (city) with the support of only one reference static population distribution sample and auxiliary data (\emph{i.e., } POI distribution). To solve this one-shot transfer learning problem, there are still several challenges:
\begin{itemize}[leftmargin=*,partopsep=0pt,topsep=1pt]
\item \emph{First, the spatial-temporal correlations between the coarse and fine-grained population distribution are complicated}. In the source data-sufficient domain, we could train a model with both coarse and fine-grained population data. However, the correlations between two distributions are spatially affected by the urban structure and they change in different locations. Thus, effectively learn transferable spatial-temporal correlations is challenging.
\item \emph{Second, it's hard to utilize the target domain's scarce reference data to guide knowledge transfer.} Single reference fine-grained population ground truth and POI distribution in the target domain could guide the knowledge transfer from the source domain. However, following experiments will prove that straightforward methods, including employing single reference fine-grained distribution as the ground truth to fine-tune model which is pre-trained in a single source domain or meta-pre-trained in multiple source domains, fail to perform well to adapt the model to the target domain. Considering the differences of large-scale structures between source and target domains, a more effective method is required for domain adaptation.
\end{itemize}

Confronting these challenges, we propose a novel \textbf{P}opulation \textbf{S}uper-\textbf{R}esolution \textbf{Net}work (\OurModelNoSpace), which follows the procedure of two stages, including pre-training for knowledge extraction in the source domain and fine-tuning for knowledge transfer from source to target domain. To extract and transfer spatial-temporal knowledge, \OurModel consists of three components: 
\begin{itemize}
    \item \textbf{S}patial-\textbf{T}emporal \textbf{Net}work (\OurSRModelNoSpace) for model-based knowledge transfer, used to extract the transferable spatial-temporal correlations between coarse-grained and fine-grained population distribution.
    \item \textbf{P}opulation \textbf{G}eneration \textbf{Net}work (\OurGANModelNoSpace) for data-based knowledge transfer, designed to transfer the relationship between POI and gridded crowd flow. It will augment the single fine-grained ground truth in target domains.
    \item \textbf{P}ixel-level \textbf{A}dversarial \textbf{D}omain \textbf{A}daptation mechanism (\emph{PADA}) as optimization-based knowledge transfer, which could mitigate the domain shift in fine-tuning stage.
\end{itemize}

\emph{In the model-based transfer network} \OurSRModelNoSpace, we design a dense connection-based population mapping network to extract spatial correlations from the coarse-grained population for fine-grained population mapping. Furthermore, we design a temporal module to enhance the transferability of above population mapping network by capturing the temporal correlations of spatial features and progressively merging them into different layers of \OurSRModel.

\emph{In the data-based transfer network} \OurGANModelNoSpace, we design a generative adversarial-based model to learn the transferable correlations between POI distribution and gridded crowd flow from the source domain and generate fine-grained population distribution in the target domain. Concretely, we utilize the dynamic representation from time-enabled long short-term memory network (LSTM) as weights to reorganize the static urban POI map and generate the sequential gridded crowd flow with a residual convolution-based network. Then, we combine the single reference fine-grained ground truth with generated crowd flow to synthesize multiple fine-grained population distribution samples in the target domain to provide more ground truth for fine-tuning.

\emph{Finally, we enable the knowledge transfer from the optimization view}. We develop a pixel-level adversarial domain adaptation framework (\emph{PADA}) to adapt our model into target domains by mitigating the domain shift between different domains in the fine-tuning stage. When employing PADA for fine-tuning, except for \OurSRModelNoSpace's regular population mapping, a pixel-level discriminator is simultaneously trained to distinguish the domain of \OurSRModelNoSpace's feature maps, whereas \OurSRModel is also optimized to confuse the discriminator. With this adversarial mechanism, we could adapt \OurSRModel while ensure its feature extraction is universal for source and target domains. That would mitigate the domain shift and improve the performance of transfer.

Our contributions are summarized as follows.
\begin{itemize}[leftmargin=*,partopsep=0pt,topsep=1pt]
	\item We present the first attempt in one-shot transfer learning for fine-grained population mapping, which is of great importance to deploy population mapping on data-scarce cities. Concretely, we develop a novel framework with three-view knowledge transfer mechanisms to infer fine-grained population distribution with scarce data in the target domain.
    \item We design a model-based transfer network \OurSRModel to transfer the spatial-temporal correlations between coarse-grained and fine-grained population distribution by its parameters. Besides, we develop a data-based transfer model \OurGANModel to synthesize fine-grained ground truth in target domains and transfer the correlation between POI distribution and gridded crowd flow. Finally, based on the aforementioned components, we design an pixel-level adversarial domain adaption fine-tuning framework \emph{PADA} to reduce the domain shift in spatial-temporal knowledge transfer between source and target domains during fine-tuning optimization.
    \item We conduct extensive experiments on real-life datasets of $4$ cities to evaluate the performance of our proposed model, including the knowledge transfer across cities and granularities. Results of on four metrics in $\times 2$ and $\times 4$ tasks demonstrate that our model has significant advantages over $8$ state-of-the-art baselines.
\end{itemize}

\section{Preliminaries} \label{sec:Prel}
In this section, we first introduce the notations and then formally define the fine-grained population mapping problem in the one-shot transfer learning scenario. Following previous works~\cite{stevens2015disaggregating, liang2019urbanfm, zong2019deepdpm}, we use gridded population distribution to formulate the fine-grained population mapping problem.
\begin{mdefinition}[Gridded Population Distribution] \label{pop}
By partitioning an area into a $H\times W$ grid map, the gridded population distribution in a single time slot is defined as tensor $X \in \mathbf{R}_+^{1 \times H\times W}$ by accumulating the users visiting each grid.
The sequential population distribution with consecutive $T$ time slots in source domain and target domain are denoted as $X_\mathcal{S}\in \mathbf{R}_+^{T\times H_\mathcal{S}\times W_\mathcal{S}}$, $X_\mathcal{T}\in \mathbf{R}_+^{T\times H_\mathcal{T}\times W_\mathcal{T}}$. 
\end{mdefinition}

Under above settings, population could be mapped into grid maps of different granularities (\emph{e.g.}, $500m\times500m$ or $2km\times2km$). Coarse-grained population distribution, with grid size $2km\times2km$ is denoted by $X^c\in \mathbf{R}_+^{T\times H\times W}$. Fine-grained population distribution \emph{e.g.}, with grid size $500m\times500m$ is denoted by $X^f\in \mathbf{R}_+^{T\times nH\times nW}, (n=2000/500=4)$. We note that both coarse-grained and fine-grained population are relative and task-specified, which will be introduced before each comparison in Experiments~\ref{sec:Exp}. In this research, $1$ time slot always contains $30$ minutes. The fine-grained population mapping task needs to recover the fine-grained distribution from coarse-grained distribution, which is formally defined as below:

\begin{problem}[Fine-grained Population Mapping]
Given the coarse-grained population distribution sequence $X^c\in \mathbf{R}_+^{T\times H\times W}$  of $T$ time slots (\emph{e.g.}, from 06:00PM to 09:00PM), estimate the fine-grained population distribution of the newest ($T$th) time slot $X^f\in \mathbf{R}_+^{1\times nH\times nW}$  (\emph{e.g.}, at 09:00PM). We note that $n$ is the upscale factor, which means population in each grid need to be partitioned into $n\times n$ sub-grids.
\end{problem}

While fine-grained population data is difficult to obtain for the majority of cities in practice, we attempt to transfer knowledge from data-sufficient city to data-scarce city with the support of single reference static fine-grained population and POI distribution in target city. The single reference static fine-grained population is available via remote sensing~\cite{stevens2015disaggregating} or questionnaire. It is critical to indicate large-scale urban structure and patterns of population in the data-scarce target domain, which is denoted by $X_{ref}^f \in \mathbf{R}_+^{1 \times nH \times nW}$. Moreover, POIs distribution characterizes the function of regions. It is also considered as a reliable and informative proxy of human activity~\cite{yuan2012discovering, Xu2016ContextawareRP, Dong2019PredictingNS,shao2021deepflowgen} in target domain. With partitioning an area into a grid map, the number of POIs of each category is defined as a tensor $P \in \mathbf{R}_+^{C\times nH\times nW}$  by accumulating POIs in each grid into $C$ categories.

By introducing knowledge transfer, single fine-grained population distribution and POI distribution for target city, we transform the fine-grained population mapping task into a one-shot transfer learning problem, since there is only one fine-grained population distribution as ground truth in target domain. This problem is formally defined as follows:
\begin{problem}[One-shot Transfer Learning For Fine-grained Population Mapping]
Given:
\begin{itemize}
    \item Sufficient coarse and fine-grained population distribution $X_\mathcal{S}^c\in \mathbf{R}_+^{L\times T\times H_\mathcal{S}\times W_\mathcal{S}}$, $X_\mathcal{S}^f\in \mathbf{R}_+^{L \times 1\times nH_\mathcal{S}\times nW_\mathcal{S}}$ of source domain $\mathcal{S}$, where $L >> 1$ is the number of samples.
    \item Sufficient coarse-grained population distribution $X_\mathcal{T}^c\in \mathbf{R}_+^{L\times T \times H_\mathcal{T}\times W_\mathcal{T}}$, one static reference fine-grained population distribution sample $X_{\mathcal{T}, ref}^f\in \mathbf{R}_+^{1 \times 1\times nH_\mathcal{T}\times nW_\mathcal{T}}$ in target domain $\mathcal{T}$.
    \item Fine-grained POI distributions (static) $P_\mathcal{S} \in \mathbf{R}_+^{C\times nH_\mathcal{S}\times nW_\mathcal{S}}$, $P_\mathcal{T} \in \mathbf{R}_+^{C\times nH_\mathcal{T}\times nW_\mathcal{T}}$ in source and target domains.
\end{itemize}
Estimate the fine-grained population distribution $X_\mathcal{T}^f\in \mathbf{R}_+^{L \times 1 \times nH_\mathcal{T}\times nW_\mathcal{T}}$ in target domain $\mathcal{T}$.
\end{problem}
\section{Methods} \label{sec:Method}
\begin{figure}[htbp]
    \centering
    \vspace{-0.3cm}
    \includegraphics[width=0.48\textwidth]{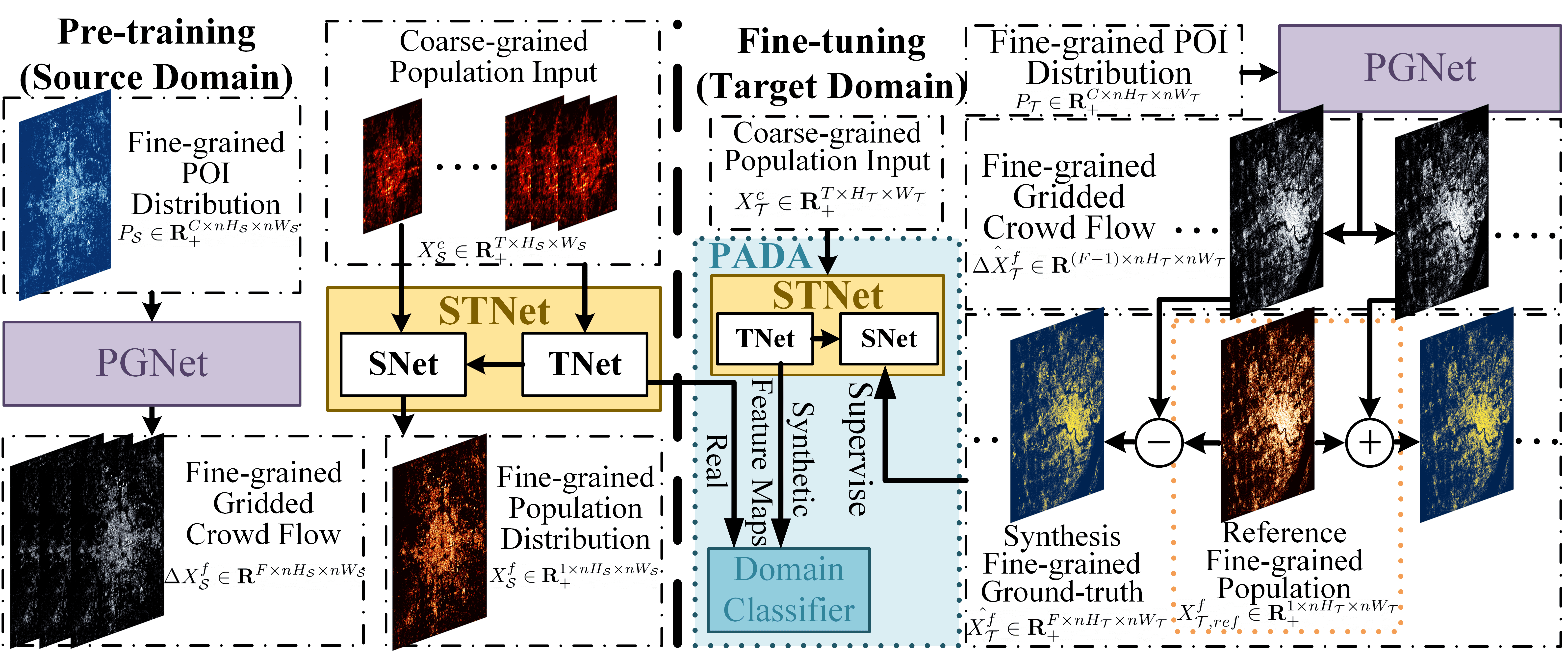}
    \vspace{-0.5cm}
    \caption{The framework of \OurModelNoSpace.}
	\vspace{-0.3cm}
    \label{fig:framework}
\end{figure}

To solve the fine-grained population mapping problem in the one-shot transfer learning scenario, we propose a novel model \OurModelNoSpace, whose basic procedure is presented in Figure~\ref{fig:framework}. \OurModel consists of three components: \OurSRModel for model-based knowledge transfer, \OurGANModel for data-based transfer, and pixel-level adversarial domain adaptation (\emph{PADA}) for optimization-based knowledge transfer. Firstly, Our \OurSRModel is designed to complete the population mapping task by modeling the complicated spatial-temporal correlations between coarse and fine-grained population distributions. Further, it is enhanced by temporal modeling network \OurTSRModelNoSpace. Secondly, \OurGANModel is designed to generate gridded crowd flow and synthesize fine-grained population distribution ground truth by capturing the spatial-temporal correlations between gridded crowd flow and POI distribution via a generative adversarial network (GAN). Finally, with the combination of \OurSRModel and \OurGANModelNoSpace, we propose \emph{PADA} for optimization-based knowledge transfer, which encourages the model to transfer the spatial-temporal knowledge while mitigating the domain shift between different domains. In this way, our model \OurModel succeeds in one-shot transfer learning for the fine-grained population mapping problem.

The training procedure of \OurModel is described as follows:
\begin{itemize}
    \item \textbf{Pre-training}: For \OurSRModelNoSpace, We employ sufficient data in source domain to infer fine-grained population distribution by sequential coarse-grained population distribution. We also train \OurGANModel to synthesize fine-grained gridded crowd flow by POI distribution in the source domain.
    \item \textbf{Fine-tuning}: First, we employ \OurGANModelNoSpace, which is pre-trained in source domain, to generate gridded crowd flow with POI distribution in target domain. Second, we combine the single reference fine-grained population distribution and the generated gridded crowd flow to synthesize fine-grained population distribution in target domain. Third, we employ the synthetic fine-grained population distribution as ground truth to support the \emph{PADA} mechanism to adapt the \OurSRModel into target domain.
\end{itemize}

\subsection{\emph{\OurSRModelNoSpace}: Spatial-Temporal Correlation Modeling}
\begin{figure}[htb]
	\centering
    \vspace{-0.3cm}
	\includegraphics[width=0.38\textwidth]{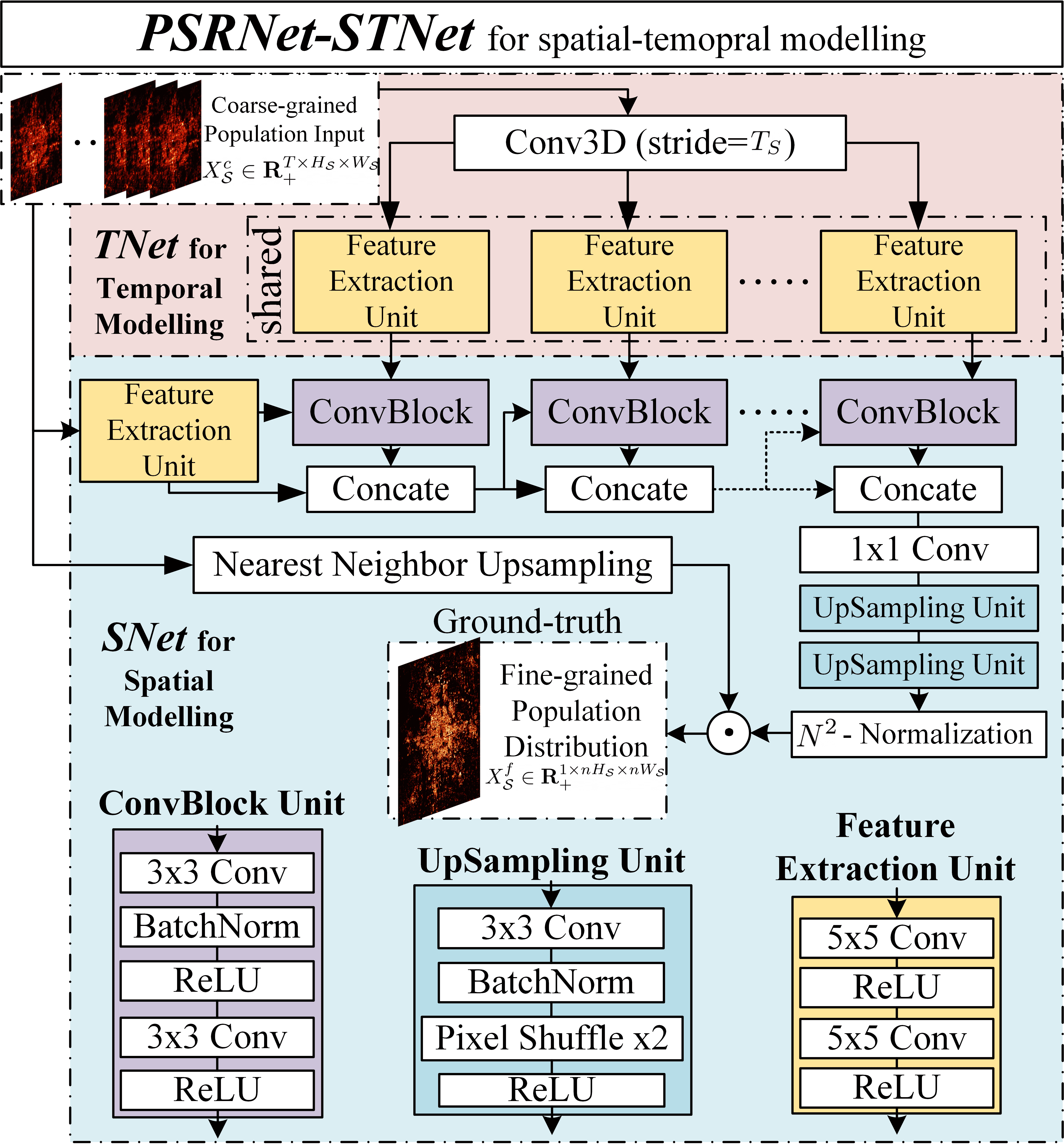}
    \vspace{-0.3cm}
    \caption{The framework of \OurSRModel in \OurModelNoSpace.}
    \vspace{-0.3cm}
    \label{fig:ST}
\end{figure}

\OurSRModel is designed to extract universal spatial-temporal correlations from the coarse-grained population input $X^c\in \mathbf{R}_+^{T\times  H\times W}$ and produce the fine-grained distribution $X^f\in \mathbf{R}_+^{1\times nH\times nW}$. As Figure~\ref{fig:ST} shows, it consists of two components: the first part is the backbone network \OurBaseSRModelNoSpace, which is for the spatial modeling of single input of coarse-grained population $X^c\in \mathbf{R}_+^{1\times H\times W}$ in the $T$th time slot; the second part is the temporal enhancement network \OurTSRModelNoSpace, which is designed to enhance \OurBaseSRModel by modeling the temporal correlation of the sequential coarse-grained population input $X^c\in \mathbf{R}_+^{T\times H\times W}$ . Now, we discuss the details of these two networks.

\subsubsection{SNet for Spatial Modeling}
We first introduce the backbone network \OurBaseSRModel for spatial modeling. As Figure~\ref{fig:ST} shows, \OurBaseSRModel can be divided into three parts: 1) the feature extraction unit for input preprocessing and preliminary feature extraction from the coarse-grained population; 2) stacked conv-blocks for advanced spatial feature extraction; and 3) the upsampling components to produce the fine-grained population map based on the feature map from previous feature extractors. We design two types of feature extractors. As shown at the bottom of Figure~\ref{fig:ST}, the preliminary feature extraction unit is made up of two $5\times 5$ convolution units which are activated by a ReLU function. Here, we choose $5\times 5$ filter to expand the receptive field of feature pre-processing. Following the preliminary feature extraction unit, we stack several dense connected conv-blocks as the advanced feature extractor to extract and fuse features again. The detailed design of conv-block unit is presented at the bottom of Figure~\ref{fig:ST}. It consists of a two-layer $3\times 3$ convolution unit activated by ReLU function and a batch-norm layer after the first $3\times 3$ convolution layer. Based on this basic unit, we apply the dense connection to construct the conv-block.

After merging all the output features from two levels of feature extractors with a $1\times 1$ convolution layer, we build an up-sampling unit to upscale the feature map. Each up-sampling unit is designed to upscale the feature map by $2$ times. One up-sampling unit consists of three layers: a $3\times 3$ convolution layer with batch-norm layer, a pixel-shuffle layer~\cite{shi2016real} with scale $2$ for rearranging and up-scaling, and a ReLU function for non-linear activation. Stacked several up-sampling units or pixel-shuffle layer of higher scale could achieve a larger up-sampling size. Different from the general image super-resolution task, the fine-grained population mapping task exhibits a specific value constraint: the population of an area equals to the total population of its sub-areas. Therefore, we finally follow $N^2$\emph{-Normalization}~\cite{liang2019urbanfm} to achieve refine the fine-grained population.

\subsubsection{TNet for Temporal Enhancement}
While \OurBaseSRModel is designed for single input of coarse-grained population $X^c\in \mathbf{R}_+^{1\times H\times W}$, a simple extension method of it for the temporal modeling is to process the sequential input by directly concatenating $T$ consecutive time slots in the channel dimension as the sequential input $X^c\in \mathbf{R}_+^{T\times H\times W}$. However, simple concatenation in the channel dimension is limited to capture this long-term correlation, which is testified in our Ablation Study~\ref{sec:Exp}. Due to the regularity and daily periodicity of dynamic population distribution, we need to consider long-term effects in the temporal modeling. Thus, we design \OurTSRModel to capture this important long-term temporal correlation, which is shown on the top of Figure~\ref{fig:ST}.

\begin{figure*}[htb]
    \centering
    \includegraphics[width=0.71\textwidth]{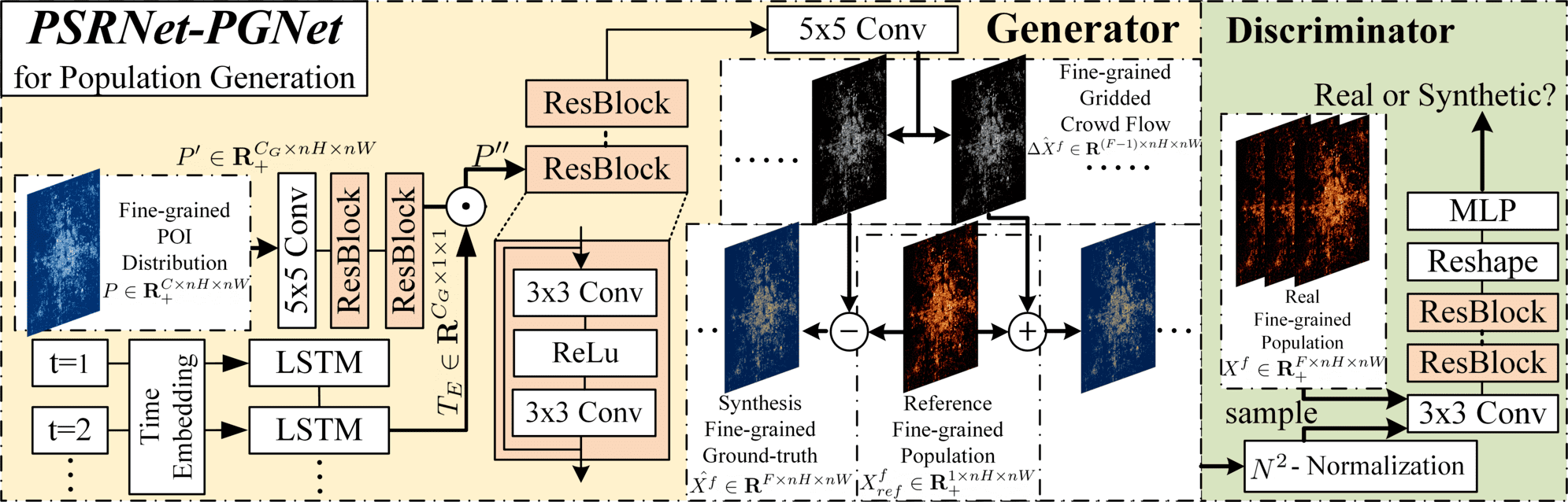}
    \vspace{-0.4cm}
    \caption{The framework of \OurGANModel in \OurModelNoSpace.}
    \vspace{-0.4cm}
    \label{fig:PG}
\end{figure*}

To model these long-term effects and avoid the limitation of simple concatenation in channel dimension, we first utilize a 3D convolution layer with stride $T_S$ to merge the coarse-grained population input in adjacent time slots. The parameter $T_S$ of stride step denotes the time window of merging features as an important factor for trading performance and model complexity. Then, we utilize the feature extraction unit with the same structure in \OurBaseSRModel to process each merged feature independently. For example, for $T_S=6$ and $T=48$, the number of the merged features is $L=48/6=8$. We use this shared feature extraction unit to process these merged features and produce new features with the same number. Finally, we progressively merge these $L$ features into $L$ conv-blocks. For example, the first feature map is fed into the first conv-block and the second map feature is fed into the second conv-block. In this way, the temporal features of $L$ different periods are merged into $L$ different layers of the spatial modeling network \OurBaseSRModelNoSpace, which can ensure that each merging operation only needs to process fewer features. It can also be regarded as an interpretative weighting scheme, more important temporal features are fed into the earlier location of the whole structure.

In summary, we design a dense connection-based \OurBaseSRModel to complete fine-grained population mapping task from the spatial view and then design \OurTSRModel to enhance the \OurBaseSRModel from the temporal view to further improve the results.

\subsection{\emph{\OurGANModelNoSpace}: Static Fine-grained Population Distribution Synthesis}
\subsubsection{Overview}
Facing the problem of lacking fine-grained population ground truth in target domain, we propose a generative model, \OurGANModelNoSpace, to generate gridded crowd flow by POI distribution. Combined with the single reference fine-grained population, \OurGANModel could synthesize multiple fine-grained population distribution. 

Population distribution, crowd flow, and POI distribution are highly associated so it's reasonable to infer one of them based on others. 
Firstly, the static reference fine-grained population distribution is strongly associated with other time slots' distribution of the same domain since they share an identical large-scale urban structure. Therefore, the static reference distribution is critical to synthesize the fine-grained population of other time slots. 
Secondly, crowd flow is defined as the difference between population distribution in consecutive time slots. Thus, with the sequence of crowd flow and static population distribution, the distribution of more time slots could be inferred by accumulating crowd flow onto static distribution iteratively.
Thirdly, POI distribution~\cite{PoI2020} characterizes the function of regions and it has been regarded as a reliable proxy for human activities in many previous works~\cite{yuan2012discovering, Xu2016ContextawareRP, Dong2019PredictingNS,shao2021deepflowgen}. Thus it's strongly associated with crowd flow or population.

Given the close relationship between POI, population, and crowd flow, our \OurGANModel learns transferable universal knowledge of the correlations between POI distribution and crowd flow in the source domain and generates fine-grained crowd flow in target domain. Then, we synthesize the fine-grained population of more time slots by accumulating the generated crowd flow onto single fine-grained reference population distribution. These synthetic population distribution data contain the knowledge of the target domain's static reference distribution, target domain's POI distribution. Finally, these synthetic population distributions will be employed as the ground truth in fine-tuning. The whole generative framework of \OurGANModel is presented in Figure~\ref{fig:PG}, which contains a generator to produce dynamic fine-grained population distribution from time-weighted POI density and a discriminator to produce learning signals for the generator via distinguishing whether the input distribution is synthetic or real.

\subsubsection{Detailed Designs}
We first introduce the design of the generator, which is at the left of Figure~\ref{fig:PG}. As the goal of human activities to move to different regions, POI describes urban function and urban structure to a large extent~\cite{yuan2012discovering}. Here, we try to generate the fine-grained crowd flow from dynamic weighted fine-grained POI density map. First, to generate the fine-grained crowd flow in time $t$ of the day, we use a learnable time embedding table to convert time $t$ into a dense feature vector. Then we feed this time vectors into a LSTM to produce the sequential representation $T_E \in \mathbf{R}^{C_G \times 1 \times 1}$ of time $t$. On the other hand, given the fine-grained POI distribution $P \in \mathbf{R}_+^{C\times nH\times nW}$, we utilize a $5\times 5$ convolution layer and $2$ res-blocks to pre-process it and obtain its feature map $P' \in \mathbf{R}^{C_G \times nH \times nW}$. Then, we multiply time embedding onto each pixel of $P'$ to obtain a time-aware feature map $P''$. Moreover, we stack $4$ res-blocks and a $5\times 5$ convolution layer to further process it to generate the fine-grained crowd flow map $\Delta X^f[t] \in \mathbf{R}^{1\times nH\times nW}$ of time $t$. 
Finally, we generate the population distributions of more time slots by adding generated crowd flow $\hat{\Delta X^f}[t]$ onto static reference fine-grained distribution $X_{ref}^f $ and employ a $N^2$-Normalization layer to regularize these population maps.
The whole generator is formulated as follows:
\vspace{-0.2cm}
\begin{gather}
\Delta X^f[t] = X^f[t + 1] - X^f[t], \\
\hat{\Delta X^f}[t]=\mathcal{F}_2(Reshape(LSTM(\mathcal{F}_E(t)))*\mathcal{F}_1(P)),\\
\hat{X^f}[l+1] = Norm(X^f_{ref} + \Sigma_{t=0}^{l}\hat{\Delta X^f}[t]),\\
\hat{X^f}[-l-1] = Norm(X^f_{ref} - \Sigma_{t=-1}^{-l-1}\hat{\Delta X^f}[t])
\end{gather}
\vspace{-0.5cm}

We first define fine-grained gridded crowd flow $\Delta X^f[t]$ as the difference fine-grained population distribution between two consecutive time slots $[t+1]$ and $[t]$. $\mathcal{F}_E$ represents the embedding layer, $Reshape$ denotes the vector extension operation, $\mathcal{F}_1$ and $\mathcal{F}_2$ denotes the two feature extraction module including convolution layers and res-blocks, $P\in \mathbf{R}_+^{C\times H\times W}$ denotes the category-aware POI distribution, $Norm$ stands for the $N^2$-Normalization layer, $\hat{\Delta X^f}[t]$ means the estimated fine-grained crowd flow in time $t$, static reference fine-grained population distribution is $X^f_{ref}$,  $\hat{X^f}[l+1]$ and $\hat{X^f}[-l-1]$ represent the synthetic fine-grained population distribution in forward and backward directions. The sequence of fine-grained population distribution $[\hat{X^f}[-l-1] \cdots \hat{X^f}[-1], X^f_{ref}, \hat{X^f}[1] \cdots \hat{X^f}[l+1]]=\hat{X^f}\in \mathbf{R}_+^{F\times nH \times nW}$ is the final output of \OurGANModelNoSpace's generator, where $F$ is the length of output sequence. 

To guide the learning of the generator, we build a discriminator to generate the learning signals. The discriminator consists of two major components: the convolution-based feature extractor which consists of a $3\times 3$ convolution unit and $3$ res-blocks and a classification module with a linear layer activated by sigmoid function. The training of \OurGANModel follows the standard GAN training procedure with a combined loss of GAN loss and MSE loss with weighting coefficient $\alpha$ as a hyper-parameter.

\subsection{Pixel-level Adversarial Domain Adaptation}\label{subsec:training}
Cooperating with the model-based transfer by \OurSRModel and data-based transfer by \OurGANModelNoSpace, we introduce the optimization-based transfer by adversarial domain adaption training mechanisms, which is presented in Figure~\ref{fig:framework}. Adversarial domain adaption~\cite{Ganin2015UnsupervisedDA, Tzeng2017AdversarialDD} is an effective transfer learning algorithm.
We adopt its basic structure while adapting it into our problem, which contains three components:
\begin{itemize}
    \item \textbf{\OurTSRModel - Feature Extractor}: We first utilize the \OurTSRModel as a universal Feature Extractor to extract feature maps in source and target domains. It is optimized to support the regression of Predictor. Whilst, it is also optimized against the Domain Classifier to confuse its classification task.
    \item \textbf{\OurBaseSRModel - Predictor}: With pre-trained \OurGANModelNoSpace, we obtain the synthetic fine-grained population data $\hat{X_\mathcal{T}^f}\in \mathbf{R}_+^{F \times nH_\mathcal{T}\times nW_\mathcal{T}}$ in target city $\mathcal{T}$ as ground truth. With the synthetic multiple fine-grained population distributions as ground truth, Predictor and Feature Extractor are optimized to complete the prediction task with MSE loss.
    \item \textbf{Domain Classifier}: Domain Classifier accepts the concatenation of feature maps from source and target domains. The classifier is highly similar to \OurGANModel but we remove the last $3$ layers and employ multi-layer perceptron (MLP) to directly classify the domain of each pixel, which contains multiple channels. Domain Classifier is optimized to classify the domain of input feature maps by working with Predictor parallelly. 
\end{itemize}
We repeat the optimization until convergence. Finally, we get a well-trained \OurSRModel to complete the fine-grained population mapping task on the target domain by learning the universal spatial-temporal knowledge between cities from the model-based, data-based, and optimization-based transfer views. 
\section{Experiments} \label{sec:Exp}

\subsection{Dataset}
We employ real-life datasets from $4$ cities, which are represented by CITY1 - CITY4, to evaluate the performance of models. These datasets are collected from mobile devices by a popular mobile localization service provider in China, which is dense in the population level and thus close to the real population distribution. It covers $4$ cities with a duration of $1$ month (2018.08$\sim$2018.09). It records the locations whenever users request localization services in the applications. 
To obtain the fine-grained gridded population distribution, each location record is projected into a grid in $500m \times 500m$ chessboard as the finest granularity, while timestamp is projected into time windows of $30$ minutes. Records are aggregated by counting the population value of each grid in each time window. We noted that the raw data with anonymous individual information is not available and we could only access the aggregated population data.
 
We also collect POI data in these cities from the public website to support the experiments. For each city, we collect about $1$ million POI instances and calculate the category-based POI distribution map, which would be used in \OurModelNoSpace. These POIs are classified into $14$ categories, including food, hotel, culture, sports, shopping, factory, recreation, institution, medical care, scenic spot, education, landmark, residence, travel \& transport, business affairs, and life service.

\subsection{Baselines}
To evaluate the performance of our model, we compare it with $8$ state-of-the-art baselines, including $2$ traditional methods (Bicubic and LightGBM), $2$ super-resolution based methods for fine-grained population mapping task (DeepDPM and UrbanFM), and $4$ advanced methods for image and video super-resolution (RCAN, DBPN, RRN, and RBPN).
\begin{itemize}[leftmargin=*,partopsep=0pt,topsep=0pt]
    \item \textbf{Bicubic~\cite{Gonzlez1981DigitalIP}: } A widely used up-sampling method for image processing, we use it to up-sample the coarse-grained population distribution.
    \item \textbf{LightGBM~\cite{ke2017lightgbm}: } It is a gradient boost regression tree-based ensemble semi-automated machine learning method, which is highly efficient and effective.
    \item \textbf{DeepDPM~\cite{zong2019deepdpm}: } It contains a CNN-based spatial mapping network and an RNN-based temporal smoothness network to extract spatial-temporal features and achieves promising results for population mapping.
    \item \textbf{UrbanFM~\cite{liang2019urbanfm}: } With its ResNet-based super-resolution network and $N^2$\emph{-Normalization} layer, UrbanFM achieves state-of-the-art performance on the fine-grained urban flow inference task.
    \item \textbf{RCAN~\cite{zhang2018rcan}: } It improves its performance of image super-resolution by considering the residual connection and channel attention in the model.
    \item \textbf{DBPN~\cite{DBPN2018}: } With back-projection units and dense connection, it repetitively up-samples and down-samples the feature maps and concatenates them for high-resolution image reconstruction.
    \item \textbf{RRN~\cite{isobe2020RRN}: } By processing the feature map with its residual module recurrently, RRN is capable to complete the video super-resolution task.
    \item \textbf{RBPN~\cite{RBPN2019}: } As the state-of-the-art method for video super-resolution, it is an enhanced version of DBPN by considering the temporal correlation with a multiple projection mechanism.
\end{itemize}

\begin{table*}[t]
\resizebox{0.95\textwidth}{!}{
\begin{tabular}{l|cccc|cccc|cccc|cccc}
\toprule
\textbf{Dataset} 
& \multicolumn{4}{c|}{\textbf{CITY2 ($\times$2)}} 
& \multicolumn{4}{c|}{\textbf{CITY3 ($\times$2)}} 
& \multicolumn{4}{c|}{\textbf{CITY2 ($\times$4)}} 
& \multicolumn{4}{c}{\textbf{CITY3 ($\times$4)}}\\ \hline
\textbf{Metrics}     & \textbf{RMSE}   & \textbf{MAE}    & \textbf{MAPE}   & \textbf{Corr}  & \textbf{RMSE}   & \textbf{MAE}    & \textbf{MAPE}   & \textbf{Corr}  & \textbf{RMSE}   & \textbf{MAE}    & \textbf{MAPE}   & \textbf{Corr}   & \textbf{RMSE}   & \textbf{MAE}    & \textbf{MAPE}   & \textbf{Corr}   \\ \midrule
\textbf{Bibubic}     & 19.463          & 12.973          & 0.325           & 0.874          & 37.504          & 20.454          & 0.457           & 0.850          & 25.879          & 17.849          & 0.447           & 0.759           & 50.437          & 28.978          & 0.648           & 0.703           \\
\textbf{LightGBM~\cite{ke2017lightgbm}}    & 19.507          & 12.755          & 0.320           & 0.883          & 37.453          & 20.021          & 0.448           & 0.851          & 26.519          & 17.849          & 0.447           & 0.768           & 50.460          & 28.654          & 0.641           & 0.707           \\ \hline
\textbf{DeepDPM~\cite{zong2019deepdpm}}     & 19.904          & 13.737          & 0.344           & 0.877          & 32.135          & 17.234          & 0.385           & 0.900          & 25.913          & 16.932          & 0.424           & 0.797           & 41.501          & 21.184          & 0.474           & 0.842           \\
\textbf{UrbanFM~\cite{liang2019urbanfm}}     & {\ul 16.546}    & {\ul 10.640}    & {\ul 0.267}     & {\ul 0.917}    & {\ul 19.808}    & {\ul 10.594}    & {\ul 0.237}     & {\ul 0.958}    & {\ul 20.900}    & {\ul 13.499}    & {\ul 0.338}     & {\ul 0.866}     & {\ul 27.722}    & {\ul 13.654}    & {\ul 0.305}     & 0.925           \\ \hline
\textbf{RCAN~\cite{zhang2018rcan}}        & 17.380          & 10.880          & 0.273           & 0.911          & 33.602          & 19.251          & 0.431           & 0.930          & 21.688          & 14.352          & 0.360           & 0.853           & 28.752          & 16.576          & 0.371           & {\ul 0.927}     \\
\textbf{DBPN~\cite{DBPN2018}}        & 17.745          & 11.404          & 0.286           & 0.908          & 20.438          & 11.355          & 0.254           & 0.956          & 25.223          & 16.128          & 0.404           & 0.818           & 29.920          & 16.965          & 0.379           & 0.910           \\
\textbf{RRN~\cite{isobe2020RRN}}& 17.836 & 11.502 & 0.288 & 0.905 & 34.074 & 19.506 & 0.436 & 0.900 & 38.977 & 25.185 & 0.631 & 0.696 & 50.008 & 31.832 & 0.712 & 0.798 \\ \
\textbf{RBPN~\cite{RBPN2019}} & 17.909 & 12.142 & 0.304 & 0.901 & 23.096 & 14.587 & 0.326 & 0.946 & 22.857 & 15.612 & 0.391 & 0.835 & 31.628 & 19.196 & 0.429 & 0.902 \\ \hline
\textbf{OurBest}     & \textbf{14.157} & \textbf{8.397}  & \textbf{0.210}  & \textbf{0.942} & \textbf{16.174} & \textbf{8.247}  & \textbf{0.184}  & \textbf{0.972} & \textbf{16.746} & \textbf{10.214} & \textbf{0.256}  & \textbf{0.917}  & \textbf{20.861} & \textbf{10.280} & \textbf{0.230}  & \textbf{0.952}  \\
\textbf{Improv.} & \textbf{14.4\%} & \textbf{21.1\%} & \textbf{21.1\%} & \textbf{7.4\%} & \textbf{18.3\%} & \textbf{22.2\%} & \textbf{22.2\%} & \textbf{8.0\%} & \textbf{19.9\%} & \textbf{24.3\%} & \textbf{24.3\%} & \textbf{31.7\%} & \textbf{24.7\%} & \textbf{24.7\%} & \textbf{24.7\%} & \textbf{19.3\%} \\ \bottomrule
\end{tabular}}
\caption{Results of our model and baselines. \textbf{Bold} denotes best (lowest) results. {\ul underline} denotes the second-best results.}
\vspace{-0.8cm}
\label{table:performance}
\end{table*}

\subsection{Experimental Settings}
In the pre-processing, $\times n$ task means with a sequence of fine-grained population distribution of shape $T \times nH \times nW$ in a certain dataset, we add the population of adjacent $n \times n$ grids together and get a sequence of coarse-grained population maps with shape $T \times H \times W$. Then we use $T \times H \times W$ distribution to infer $T \times nH \times nW$ distribution. Fine-grained POI distribution and single reference fine-grained population distribution are always of fine-grained shape $C \times nH \times nW$ and $1\times nH \times nW$.

For source cities, we randomly select $70\%$ time slots as the training set and utilize the remaining $15\%$ and $15\%$ time slots as validation set and test set. For target cities, we use $1$ week as the test set to evaluate the performance on the fine-grained population mapping task in the transfer learning scenario, while its first time slots is used as the reference fine-grained population distribution in target domain, it can be regarded as a one-shot transfer learning scenario.

The default settings of \OurSRModel are length of population sequence $T=48$, time stride $T_S=6$, number of layers $L=48/6=8$, base input/output channel $C_B=64$, time channel $C_T=16$. For the generator of \OurGANModelNoSpace, we use base channel $C_G=64$. For the discriminator of \OurGANModelNoSpace, we use number of layers $L=3$, spatial stride $S_S=4$, base channel $C_D=1$.

We evaluate the model with $4$ metrics: Root Mean Square Error (RMSE), Mean Absolute Error (MAE), Mean Absolute Percentage Error (MAPE), and Pearson Correlation Coefficient (Corr). Based on these metrics, we calculate the error between estimated fine-grained population distributions and their ground truth.

We conducted experiments on Ubuntu 18.04.3 LTS system with 4x NVIDIA GTX 2080Ti using Python 3.6.10 and PyTorch 1.6.0. Our models, experiment code, and datasets are available via \href{https://github.com/erzhuoshao/PSRNet-CIKM}{\textcolor{blue}{Github}}.

\subsection{One-shot Transfer across Cities}
To verify the effectiveness of our \OurModelNoSpace, we compare our model with baselines on the $\times 2$ and $\times 4$ fine-grained population mapping task in the cross-cities scenarios.

For all baselines, we firstly pre-train their models in CITY1 with sufficient data and employ MAML~\cite{finn2017MAML} on all cities except the target city as an additional meta-pre-training. Then we use the single reference fine-grained population distribution to fine-tune them on the target city (CITY2 or CITY3) and obtain the final mapping results.

In this subsection, to train our proposed \OurModelNoSpace, firstly, we employ CITY1 to pre-train \OurSRModel and \OurGANModelNoSpace. Secondly, we employ the pre-trained \OurGANModel to synthesize distribution in more time slots with the single reference fine-grained population distribution. Finally, we employ \emph{PADA}~\ref{subsec:training} to fine-tune the pre-trained \OurSRModel with the synthetic population distribution. Then, \OurModel is enabled to generate fine-grained population distribution. 
We note that our \OurModel is not only compared with the architecture of baselines but also compared with MAML~\cite{finn2017MAML}, which is the meta-learning mechanism for knowledge transfer across cities. We note that \OurModel only utilize $1$ source city (CITY1) for pre-training while $3$ source cities are used to support baselines' meta-learning. That would bring advantages for baselines in this comparison.

Table~\ref{table:performance} shows the performance of all baselines and our \OurModelNoSpace, where the notation Improv. indicates the percentage of reduction of RMSE, MAE, MAPE, and the increase of Corr of our method when compared with the best baselines. From these results, we can find that \OurModel always outperforms all baselines in all metrics. Although UrbanFM, DBPN, RRN, and RBPN reach comparable results in some scenarios, none of them could outperform \OurModel in any task. They obtain $16,546$, $19.8083$, $20.8997$, and $27.7215$ by RMSE, which are the second-best results in $\times2$ and $\times4$ tasks when CITY2 and CITY3 are target cities. Compared with these results, \OurModel reduces the RMSE by $14.4\%$, $18.3\%$, $19.9\%$, and $24.7\%$, while other baselines could only achieve comparable performance in minor metrics or tasks, and fail to sustain a consistently comparable performance in other tasks. UrbanFM is the only comparable baseline since is specially designed for urban population mapping scenario, while it is still less competitive than \OurModelNoSpace.

\begin{figure}[htbp]
    \centering
    \vspace{-0.2cm}
    \subfigure[Ground truth.]{
        \includegraphics*[width=0.10\textwidth]{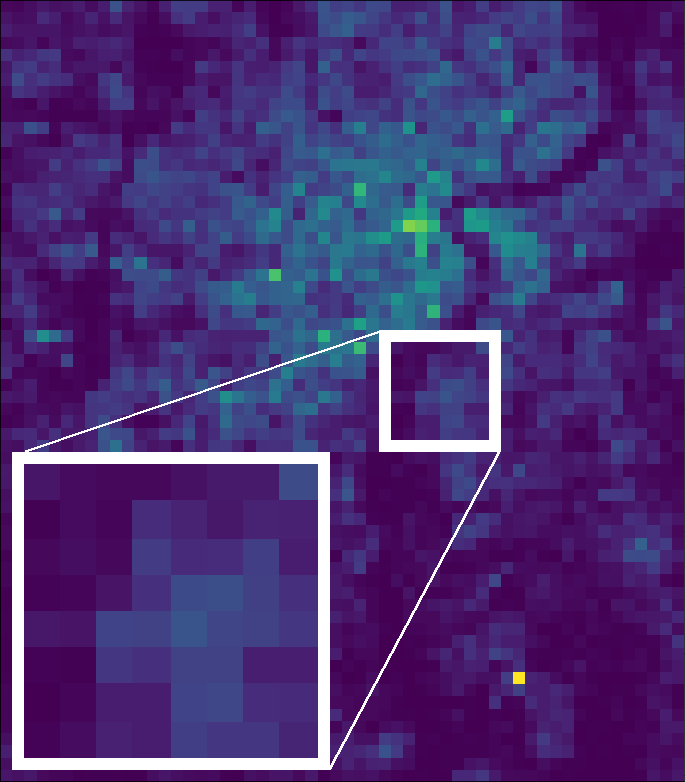}}
    \subfigure[Our model.]{
        \includegraphics*[width=0.10\textwidth]{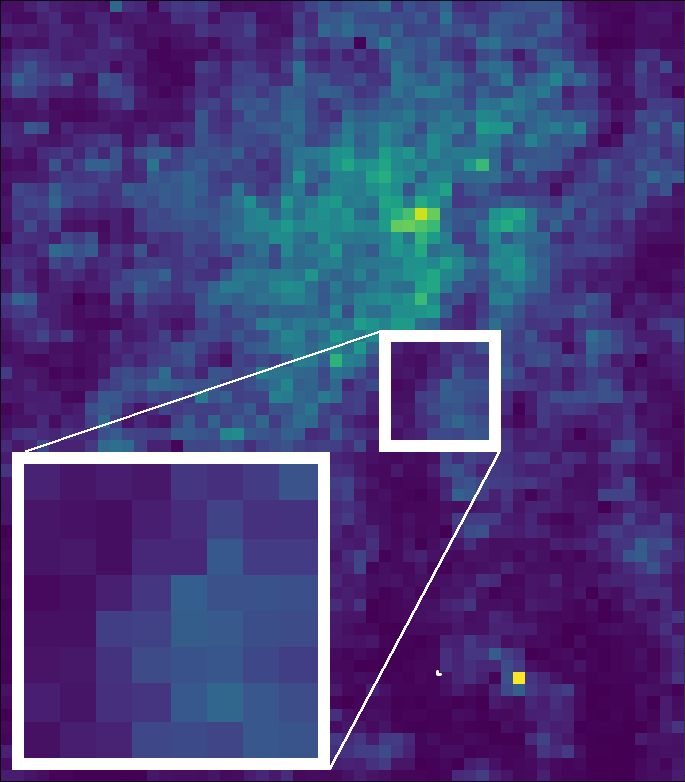}}
    \subfigure[UrbanFM.]{
        \includegraphics*[width=0.10\textwidth]{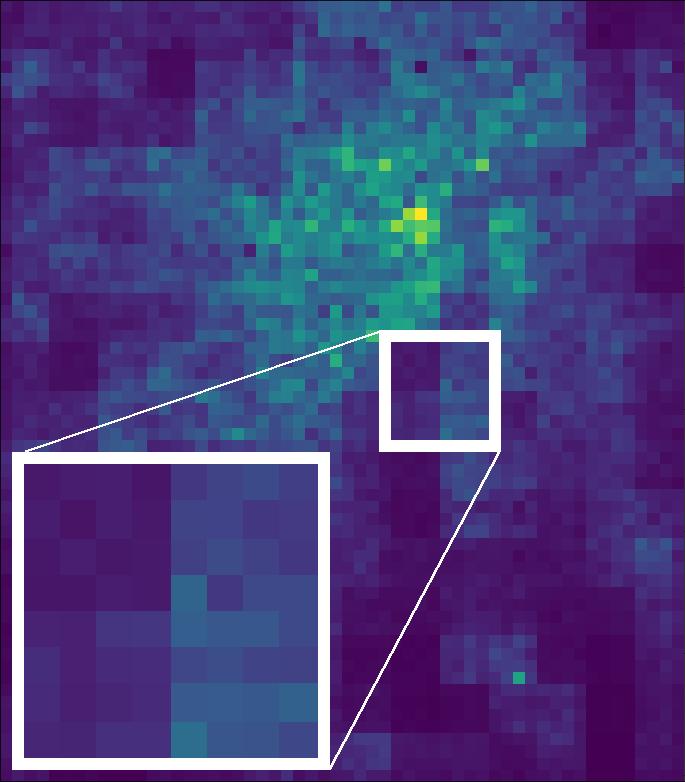}}
    \subfigure[RBPN.]{
        \includegraphics*[width=0.10\textwidth]{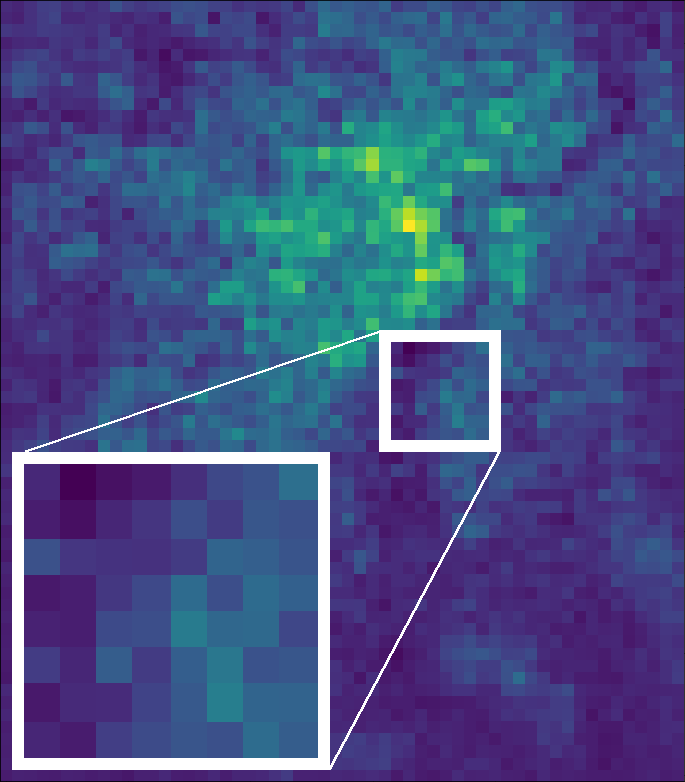}}
    \vspace{-0.6cm}
    \caption{Visualization of $\times 4$ results in CITY2.}
    \vspace{-0.2cm}
    \label{fig:Visualization}
\end{figure}
In addition to the numerical evaluation of models' performance, we also compare them by visualization. The ground truth map and predicted fine-grained population maps at 09:00 AM of \OurModelNoSpace, UrbanFM, and RBPN in CITY2's $\times4$ task are demonstrated in Figure~\ref{fig:Visualization}. 

In these figures, the lighter places denote greater population values and vice versa. The region circled by the smaller white rectangle is a region that contains $8 \times 8$ fine-grained grids mapped from $2 \times 2$ coarse-grained grids in $\times4$ task. The bigger rectangles are $3\times$ zooms of smaller rectangles. We note that the river crosses the city from the bottom to the right-up corner of the image and the selected region is on the bank of this river.

In Figure~\ref{fig:Visualization}, \OurModelNoSpace's distribution have the closest shapes and textures with the ground truth distribution. For example, compared with \OurModelNoSpace, UrbanFMs' predicted fine-grained population suddenly changes on the margins of different coarse-grained pixels so the zoomed area has an obvious unnatural vertical dividing line. Besides, the map of RBPN is much lighter and rougher than \OurModel in the zoomed region. These results show that compared with RBPN, \OurModel could capture the pattern of the river and produce more reasonable fine-grained distributions.

In summary, numerical comparison and visualization prove our proposed \OurModel has significant advantages over state-of-the-art baselines in fine-grained population mapping tasks in the cross-cities transfer learning scenario.

\subsection{One-shot Transfer across Granularities}
In this subsection, we research an extreme data-scarce scenario in which any data out of the target cities is unobtainable. In this scenario, only coarse-grained distribution, single reference fine-grained population distribution, and POI distribution in target cities are available. This situation brings us a new challenge that we cannot transfer knowledge from other cities. Inspired by the self-supervised zero-shot super-resolution method~\cite{shocher2018zero}, we design a novel cross-granularities self-supervision task in which we utilize coarser-grained distribution (down-sampled from coarse-grained distribution) to infer coarse-grained distribution as universal knowledge extraction.

Given the significant diversity between cities' large-scale structures, it is a natural assumption that the domain shift between the population of same city's different granularities was smaller than the population of different cities' same granularity. Concretely, instead of transferring knowledge from CITY1 $1km \rightarrow 500m$ task to CITY2 $1km \rightarrow 500m$ task (\emph{e.g.,} CITY2's $\times 2$ task), in this subsection, we are attempting to test the feasibility to transfer knowledge from CITY2 $2km \rightarrow 1km$ to CITY2 $1km \rightarrow 500m$. Therefore, we design a novel cross-granularities knowledge transfer task to test the capability of each model, in which we pre-train all baselines to infer $1km \times 1km$ population from $2km \times 2km$ population and use the single static reference $500m \times 500m$ distribution as ground truth to fine-tune these models. Finally, all baselines accept $1km \times 1km$ population to estimate the distribution of $500m \times 500m$. For \OurModelNoSpace, we adapt our proposed training procedure~\ref{subsec:training} into this scenario by pre-training both \OurSRModel and \OurGANModel to generate the population distribution of $1km \times 1km$.

\begin{table}[]
\resizebox{0.48\textwidth}{!}{
\begin{tabular}{l|cccc|cccc}
\toprule
\textbf{Dataset}     & \multicolumn{4}{c|}{\textbf{CITY2}} & \multicolumn{4}{c}{\textbf{CITY3}} \\\hline
\textbf{Granularity} & \multicolumn{4}{c|}{\textbf{(2km$\rightarrow$1km)$\rightarrow$(1km$\rightarrow$500m)}} & \multicolumn{4}{c}{\textbf{(2km$\rightarrow$1km)$\rightarrow$ (1km$\rightarrow$500m)}} \\\hline
\textbf{Metrics}  & \textbf{RMSE}           & \textbf{MAE}           & \textbf{MAPE}          & \textbf{Corr}         & \textbf{RMSE}           & \textbf{MAE}           & \textbf{MAPE}          & \textbf{Corr}         \\ \midrule
\textbf{Bibubic}  & 19.463                  & 12.973                 & 0.325                  & 0.874                 & 37.504                  & 20.454                 & 0.457                  & 0.850                 \\
\textbf{LightGBM} & 22.205                  & 14.033                 & 0.352                  & 0.852                 & 41.810                  & 21.799                 & 0.488                  & 0.839                 \\ \hline
\textbf{DeepDPM}  & 19.698                  & 13.314                 & 0.334                  & 0.873                 & 34.333                  & 18.854                 & 0.422                  & 0.879                 \\
\textbf{UrbanFM}  & {\ul 17.677}            & {\ul 11.219}           & {\ul 0.281}            & {\ul 0.906}           & {\ul 21.222}            & {\ul 11.647}           & {\ul 0.260}            & {\ul 0.953}           \\ \hline
\textbf{RCAN}     & 18.698                  & 12.693                 & 0.318                  & 0.889                 & 30.956                  & 19.144                 & 0.428                  & 0.900                 \\
\textbf{DBPN}     & 18.402                  & 12.102                 & 0.303                  & 0.898                 & 24.980                  & 14.237                 & 0.318                  & 0.936                 \\
\textbf{RRN}      & 19.195                  & 13.181                 & 0.330                  & 0.888                 & 31.045                  & 17.803                 & 0.398                  & 0.902                 \\
\textbf{RBPN}     & 21.382                  & 14.458                 & 0.362                  & 0.859                 & 35.185                  & 21.869                 & 0.489                  & 0.881                 \\ \hline
\textbf{\OurModelNoSpace}   & \textbf{13.073}         & \textbf{7.950}         & \textbf{0.199}         & \textbf{0.950}        & \textbf{16.164}         & \textbf{7.886}         & \textbf{0.176}         & \textbf{0.972}        \\ 
\textbf{Improv.}  & \textbf{26.0\%}         & \textbf{29.1\%}        & \textbf{29.1\%}        & \textbf{4.8\%}        & \textbf{23.8\%}         & \textbf{32.3\%}        & \textbf{32.3\%}        & \textbf{2.0\%} \\ \bottomrule
\end{tabular}}
\caption{Performance of our model and baselines on cross-granularities knowledge transfer task. \textbf{(2km$\rightarrow$1km)$\rightarrow$(1km$\rightarrow$500m)} means we pre-train \OurModel in source domain, in which we infer population of $1km \times 1km$ from population of $2km \times 2km$, while in target domain, we infer $500m \times 500m$ from $1km \times 1km$.}
\vspace{-0.8cm}
\label{table:granularity}
\end{table}

Table~\ref{table:granularity} demonstrate the performance of \OurModel and baselines in cross-granularities knowledge transfer scenario. According to Table~\ref{table:granularity}, our \OurModel could always achieve the best performance in both cities and all metrics, while UrbanFM always reaches the second-best results. We note that in cross-cities scenario of $\times 2$ task, \OurModelNoSpace's RMSE is $14.1569$ and $16.1735$ in CITY2 and CITY3, whereas these numbers are $13.0725$ and $15.9907$ in cross-granularities scenario, which shows the domain shift in the cross-granularities scenario is smaller than cross-cities scenario. Therefore, the knowledge of same city's coarse-grained distribution is more transferable than other cities' fine-grained distribution, which validates the aforementioned assumption. Furthermore, it also strongly implies \OurModel is potential to generate the finer-grained population distribution (\emph{i.e.,} of $250m \times 250m$ or even finer granularity) as long as we employ a finer-grained static reference distribution to fine-tune \OurModelNoSpace. Unfortunately, we are not able to validate the results without enough ground truth in finer granularity.

\begin{table*}[h]
\resizebox{0.95\textwidth}{!}{
\begin{tabular}{l|cccc|cccc|cccc|cccc}
\toprule
\textbf{Dataset} 
& \multicolumn{4}{c|}{\textbf{CITY2 ($\times$2)}}
& \multicolumn{4}{c|}{\textbf{CITY3 ($\times$2)}}
& \multicolumn{4}{c|}{\textbf{CITY2 ($\times$4)}}
& \multicolumn{4}{c}{\textbf{CITY3 ($\times$4)}}\\ \hline
\textbf{Metrics}     & \textbf{RMSE}   & \textbf{MAE}   & \textbf{MAPE}  & \textbf{Corr}  & \textbf{RMSE}   & \textbf{MAE}   & \textbf{MAPE}  & \textbf{Corr}  & \textbf{RMSE}   & \textbf{MAE}    & \textbf{MAPE}  & \textbf{Corr}  & \textbf{RMSE}   & \textbf{MAE}    & \textbf{MAPE}  & \textbf{Corr}  \\
\midrule
\textbf{UrbanFM}     & 16.546          & 10.640         & 0.267          & 0.917          & 19.808          & 10.594         & 0.237          & 0.958          & 20.900          & 13.499          & 0.338          & 0.866          & 27.722          & 13.654          & 0.305          & 0.925          \\
\textbf{\OurBaseSRModelNoSpace}        & 15.395          & 9.426          & 0.236          & 0.930          & 18.589          & 9.201          & 0.206          & 0.963          & 19.444          & 12.105          & 0.303          & 0.887          & 26.157          & 12.616          & 0.341          & 0.934          \\
\textbf{\OurBaseSRModelNoSpace+\OurGANModelNoSpace}  & 15.021          & 9.389          & 0.235          & 0.933          & 17.740          & 9.158          & 0.205          & 0.966          & 17.880          & 11.467          & 0.287          & 0.903          & 22.489          & 11.350          & 0.308          & 0.945          \\
\textbf{\OurSRModelNoSpace}       & 15.050          & 9.304          & 0.233          & 0.934          & 18.264          & 9.149          & 0.205          & 0.966          & 18.318          & 11.187          & 0.280          & 0.902          & 25.531          & 12.332          & 0.313          & 0.938          \\
\textbf{Meta-\OurSRModelNoSpace}  & 14.973          & 9.270          & 0.232          & 0.934          & 17.968          & 8.957          & 0.200          & 0.967          & 18.561          & 11.425          & 0.286          & 0.899          & 25.861          & 12.429          & 0.313          & 0.936          \\
\textbf{\OurSRModelNoSpace+\OurGANModelNoSpace} & \textbf{14.157} & \textbf{8.397} & \textbf{0.210} & \textbf{0.942} & {\ul 16.420}    & {\ul 8.451}    & {\ul 0.189}    & {\ul 0.971}    & {\ul 16.795}    & {\ul 10.336}    & {\ul 0.259}    & {\ul 0.916}    & {\ul 21.020}    & {\ul 10.392}    & {\ul 0.268}    & {\ul 0.952}    \\
\textbf{\OurModelNoSpace}      & {\ul 14.714}    & {\ul 9.066}    & {\ul 0.227}    & {\ul 0.937}    & \textbf{16.174} & \textbf{8.247} & \textbf{0.184} & \textbf{0.972} & \textbf{16.746} & \textbf{10.214} & \textbf{0.256} & \textbf{0.917} & \textbf{20.861} & \textbf{10.280} & \textbf{0.263} & \textbf{0.952} \\
\bottomrule
\end{tabular}}
\caption{Performance of different variants of our model in cross-cities knowledge transfer task.}
\vspace{-0.8cm}
\label{table:ablation}
\end{table*}

\subsection{Ablation Study}
In this section, we conduct an ablation study in cross-cities scenario to evaluate the effectiveness of each proposed component of \OurModelNoSpace. To investigate their effect, we compare different variants of our method. To evaluate \OurBaseSRModelNoSpace's capability of spatial-temporal knowledge extraction, UrbanFM is also introduced in this comparison. All the variants are introduced as follows:

\begin{itemize}[leftmargin=*]
    \item \textbf{\OurBaseSRModelNoSpace} is the backbone population mapping network of \OurModelNoSpace, which accepts $48$ coarse-grained population maps as multiple input channels.
    \item \textbf{+\OurTSRModelNoSpace} denotes that we introduce \OurTSRModel to enhance the temporal modeling of sequential input. \OurBaseSRModelNoSpace+\OurTSRModel is simplified as \OurSRModelNoSpace.
    \item \textbf{+\OurGANModelNoSpace} represents we employ \OurGANModel to generate synthetic fine-grained population map in target domain. Then these generated maps are used to fine-tune the pre-trained population mapping network.
    \item \textbf{+\emph{MAML}} means we employ meta-learning algorithm MAML~\cite{finn2017MAML} to drive an additional meta-training with three cities' data except the target city for population mapping models.
    \item \textbf{+\emph{PADA}} means employ pixel-level adversarial domain adaptation~\ref{sec:Method} to fine-tune the model in target domain. The complete version of our proposed method is \OurSRModelNoSpace+\OurGANModelNoSpace+\emph{PADA} (denoted as \OurModelNoSpace).
\end{itemize}

The results in Table~\ref{table:ablation} brings us several conclusions:
\begin{itemize}
    \item Our \OurBaseSRModel could outperform UrbanFM, which shows that \OurBaseSRModel could capture spatial patterns in population maps by its more advanced architecture.
    \item In \OurSRModelNoSpace, combined with \OurTSRModelNoSpace, which could effectively exploit temporal information, the feature maps of different time slots are fed into different layers. Therefore, each layer of \OurSRModel only needs to process less information. \OurSRModel outperforms \OurBaseSRModel indicates that \OurSRModel is better to capture transferable features in all scenarios. 
    \item By comparing the performance of \OurSRModel and MAML+\OurSRModelNoSpace, we find that although MAML slightly improves \OurSRModelNoSpace's performance in $\times 2$ tasks, it fails to sustain this improvement in $\times 4$ tasks. That shows the meta-learning method failed to transfer knowledge across cities stably.
    \item \OurBaseSRModelNoSpace+\OurGANModel outperforms \OurBaseSRModelNoSpace, while \OurSRModelNoSpace+\OurGANModel outperforms \OurSRModel in all scenarios. That shows our \OurGANModel could always improve the performance by providing more synthetic ground truth in target city and transferring the correlation between POI distribution and gridded crowd flow.
    \item Compared with \OurSRModelNoSpace+\OurGANModelNoSpace, the performance of \OurModel (\OurSRModelNoSpace+\OurGANModelNoSpace+\emph{PADA}) is better in $\times4$ task in CITY2 and all tasks in CITY3. Although \OurSRModelNoSpace+\OurGANModel performs slightly better for $\times2$ task in CITY2, \OurModel is still comparable.
\end{itemize}

In summary, our proposed model-based, data-based, and optimization-based transfer learning components in \OurModel bring performance gain in one-shot transfer learning fine-grained population mapping task. That proves the reasonability of our design.
\section{Related Works} \label{sec:Related}
\subsection{Image and Video Super-Resolution}
With the application of deep learning, research community~\cite{wang2019deep, dong2014learning, DBPN2018, zhang2018rcan, li2019fast} makes significant progress on image super-resolution task. SRCNN~\cite{dong2014learning} utilizes several convolution layers to build the first end-to-end framework for image super-resolution. With the basic idea of first building deep convolution networks to extract features and then up-sampling to obtain the high-resolution image, many following up works are proposed with more advanced network design~\cite{DBPN2018, zhang2018rcan}, specific loss function~\cite{ledig2017photo} and so on. Considering the temporal correlation between multiple frames, image super-resolution upgrades to the video super-resolution task. While some works~\cite{liao2015video} try to model spatial-temporal correlation via the motion compensation between different frames by optical flow or learning, others~\cite{RBPN2019, li2019fast} try to directly learn the spatial-temporal dependency with different sequential structures like recurrent neural networks. Different from these existing works on the general image/video super-resolution task, we focus on the fine-grained population mapping task and propose effective methods to transfer the spatial-temporal knowledge to promote the performance in cities without fine-grained data.
\subsection{Fine-Grained Population Mapping}
By applying the successful practice of image super-resolution into fine-grained population mapping, DeepDPM~\cite{zong2019deepdpm} and UrbanFM~\cite{liang2019urbanfm} are the most related work to our work. DeepDPM~\cite{zong2019deepdpm} first utilizes SRCNN~\cite{dong2014learning} with stacking structure to up-sample the static population distribution and then utilizes LSTM to refine the population variation in the temporal dimension. To infer the fine-grained crowd flow, UrbanFM~\cite{liang2019urbanfm} proposes a ResNet-based network structure with applying the recent practice from image super-resolution and also consider the effects of external factors like holidays in the model. While they achieve promising performance in the city with enough data, they require a large number of fine-grained data to train the whole model, which is not available for most of the cities. Different from them, our work considers the transferred fine-grained population mapping task and proposes to transfer the spatial-temporal knowledge from the data and model view to improve the mapping performance on these cities without fine-grained data.
\subsection{Transfer Learning Among Cities}
Knowledge transferring between cities is an important topic in urban computing. Wei et al.~\cite{Wei2016TransferKB} propose FLORAL with learning semantically related dictionaries and transferring dictionaries and instances to predict air quality in different cities. Wang et al.~\cite{Wang2019CrossCityTL} propose to use slide information from public check-ins to align regions from different cities to enable the explicit knowledge transfer in the crowd flow prediction task. Yao et al.~\cite{Yao2019LearningFM} apply MAML optimization methods to enable the multi-cities crowd flow prediction. These existing works focus on the multi-variant time series prediction problem and are not directly available for our mapping task. Furthermore, different from these works which only transfer knowledge from single view, our framework enables the spatial-temporal knowledge transfer from model, data, and optimization views. 
\section{Conclusion}
\label{sec:Conclusion}
In this paper, we investigated the fine-grained population mapping problem in the transfer learning scenario. We transfer this problem into a one-shot transfer learning problem for the population mapping task. We proposed a novel model by transferring the spatial-temporal knowledge from model view, data view, and optimization view. We designed a sequential population mapping network to capture the complicated correlation between the population of different granularities. Furthermore, we proposed a generative model to synthesize multiple fine-grained population samples in target domain with POI distribution. Finally, we utilized the adversarial adaptation methods to fine-tune the pre-trained model and transfer the universal spatial-temporal knowledge.

\begin{acks}
This work was supported in part by The National Key Research and Development Program of China under grant 2018YFB1800804, the National Nature Science Foundation of China under U1936217,  61971267, 61972223, 61941117, 61861136003, Beijing Natural Science Foundation under L182038, Beijing National Research Center for Information Science and Technology under 20031887521, and research fund of Tsinghua University - Tencent Joint Laboratory for Internet Innovation Technology.
\end{acks}

\newpage
\newpage
\bibliographystyle{ACM-Reference-Format}
\bibliography{acmart}
\end{document}